\documentclass[journal]{IEEEtran}
\ifCLASSINFOpdf
\else
\fi
\usepackage{amsmath}

\usepackage{amssymb}
\DeclareMathAlphabet{\pazocal}{OMS}{zplm}{m}{n}
\usepackage{bbm}
\usepackage{lettrine}
\usepackage{array} 

\usepackage{float}
\usepackage[export]{adjustbox}
\usepackage{graphicx}
\usepackage[colorlinks=true, allcolors=blue]{hyperref}
\usepackage[utf8]{inputenc}
\usepackage{caption}
\usepackage{subcaption}
\usepackage{multirow}
\usepackage{multicol}
\usepackage{amsmath}
\usepackage{amsfonts}
\usepackage{bm}
\usepackage{bbm}
\usepackage{algorithm}
\usepackage{algorithmicx}
\usepackage{algpseudocode}
\usepackage{booktabs}
\hypersetup{pdfstartview={XYZ null null 1.00}}

\begin{document}
%

\title{TFNet: Tuning Fork Network with Neighborhood Pixel Aggregation for Improved Building Footprint Extraction}
%
%
%

\author{Muhammad Ahmad Waseem, Muhammad Tahir, \IEEEmembership{Senior Member, IEEE}, Zubair Khalid \IEEEmembership{Senior Member, IEEE}, and Momin Uppal \IEEEmembership{Senior Member, IEEE}
\thanks{The authors are with Department of Electrical Engineering, Lahore University of Management Sciences, DHA Lahore Cantt., 54792, Lahore Pakistan. This work was supported by the Higher Education Commission (HEC) of Pakistan through a Grand Challenge Fund Grant No. GCF-521. \emph{Corresponding author: Momin Uppal, e-mail: momin.uppal@lums.edu.pk}}
}

%
%

\markboth{IEEE TRANSACTIONS ON GEOSCIENCE AND REMOTE SENSING, ~Vol.~xx, ~2023}%
{Shell \MakeLowercase{\textit{et al.}}: Bare Demo of IEEEtran.cls for IEEE Journals}
%



\maketitle

\begin{abstract}
This paper considers the problem of extracting building footprints from satellite imagery -- a task that is critical for many urban planning and decision-making applications. While recent advancements in deep learning have made great strides in automated detection of building footprints, state-of-the-art methods available in existing literature often generate erroneous results for areas with densely connected buildings. Moreover, these methods do not incorporate the context of neighborhood images during training thus generally resulting in poor performance at image boundaries. In light of these gaps, we propose a novel Tuning Fork Network (TFNet) design for deep semantic segmentation that not only performs well for widely-spaced building but also has good performance for buildings that are closely packed together. The novelty of TFNet architecture lies in a a single encoder followed by two parallel decoders to separately reconstruct the building footprint and the building edge. In addition, the TFNet design is coupled with a novel methodology of incorporating neighborhood information at the tile boundaries during the training process. This methodology further improves performance, especially at the tile boundaries. For performance comparisons, we utilize the SpaceNet2 and WHU datasets, as well as a dataset from an area in Lahore, Pakistan that captures closely connected buildings. For all three datasets, the proposed methodology is found to significantly outperform benchmark methods.
\end{abstract}


%
\IEEEpeerreviewmaketitle

\section{Introduction}\label{sec:intro}
\IEEEPARstart{T}{he} availability of building footprints is crucial for informed urban planning and decision-making. They are used by urban planners for applications that include, but are not limited to, temporal change detection, population density estimation, public services planning, infrastructure management, and disaster damage assessment. The traditional method of generating these footprints involve tedious onsite measurements and surveys, which are extremely costly as well as time consuming. But recent advancements in artificial intelligence coupled with ready access to high resolution satellite imagery has enabled development of automated toolsets that promise significant time and cost savings over traditional onsite surveys. 

\begin{figure*}
\setlength\tabcolsep{2pt}
\centering
\begin{tabular}{l|c|c|c|c}
 &
 \textbf{Satellite Image} &
 \textbf{GT Mask} &
 \textbf{SOTA Prediction \cite{jiwani2021semantic}} &
 \textbf{Our Prediction} \\
 \rotatebox{90}{Connected Buildings} &
 \includegraphics[height=0.15\linewidth, width=0.22\linewidth]{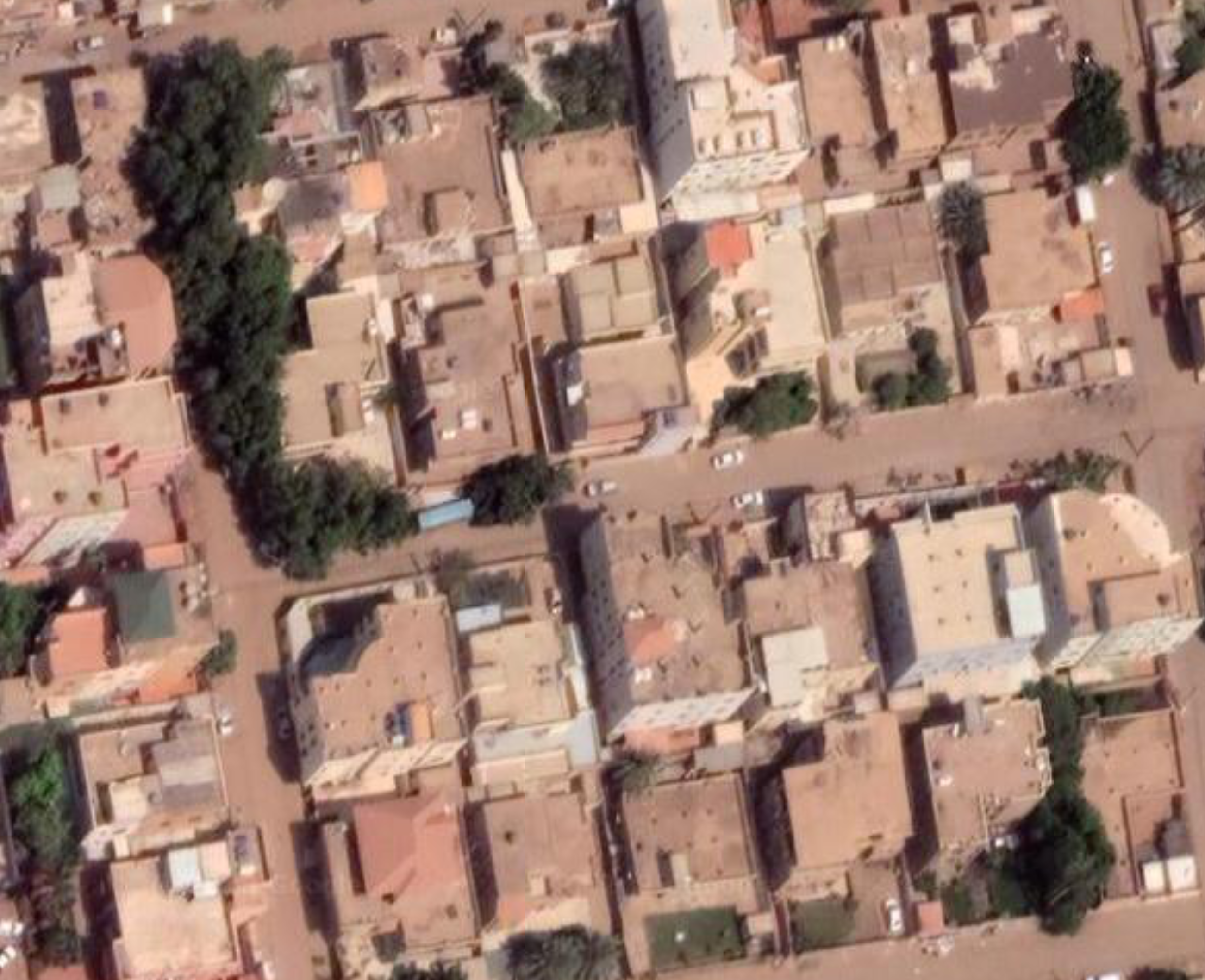} &
 \includegraphics[height=0.15\linewidth, width=0.22\linewidth]{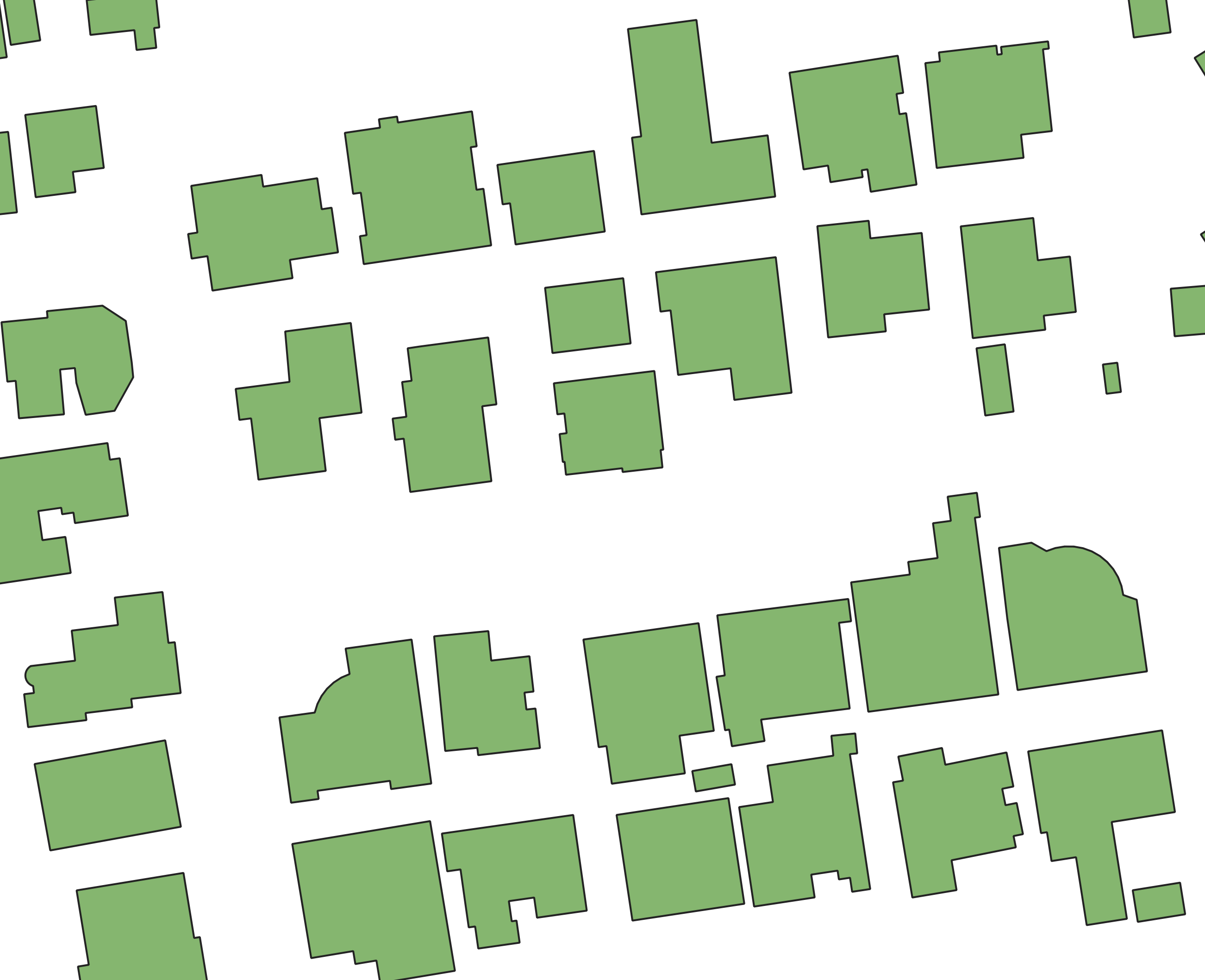} &
 \includegraphics[height=0.15\linewidth, width=0.22\linewidth]{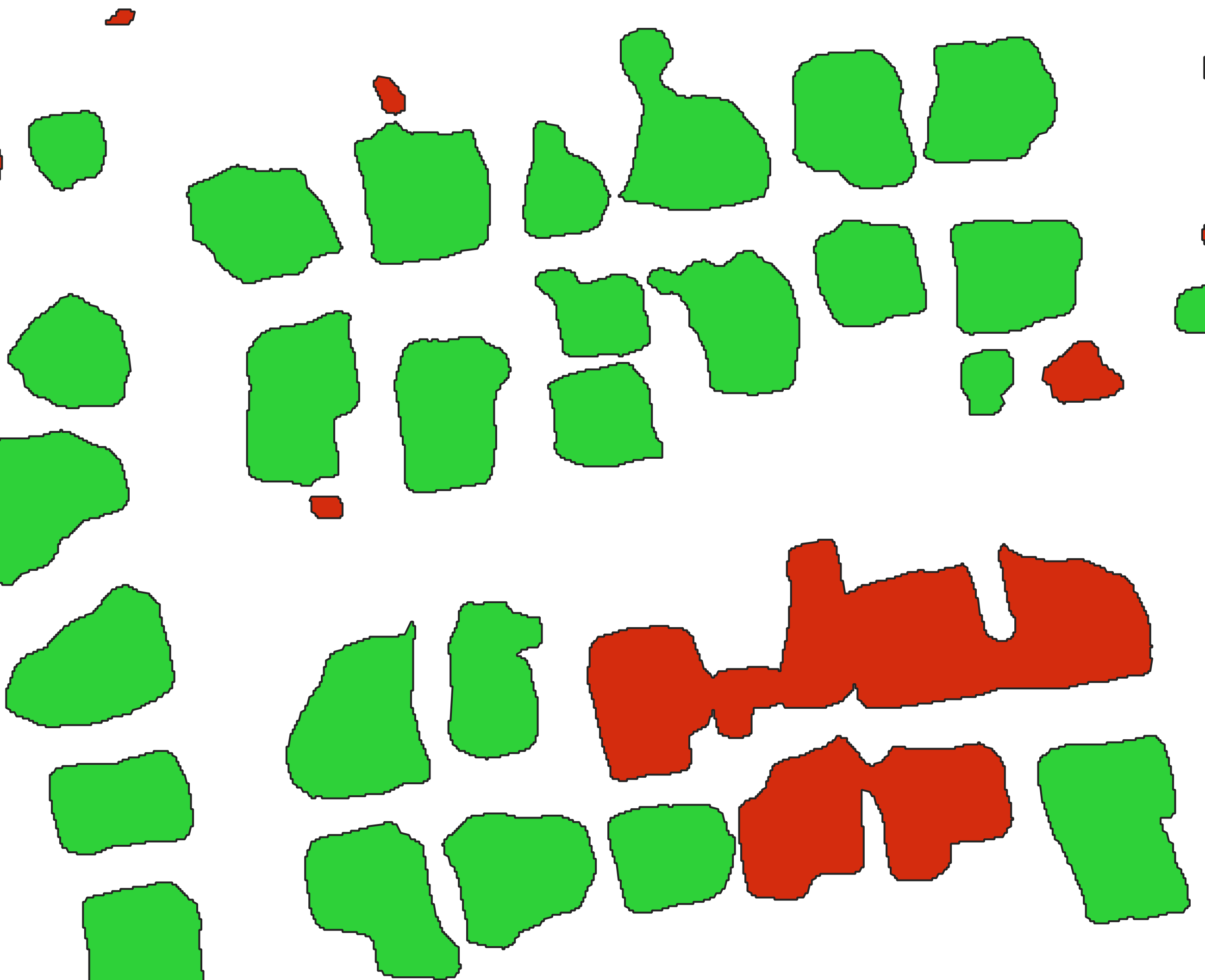} &
 \includegraphics[height=0.15\linewidth, width=0.22\linewidth]{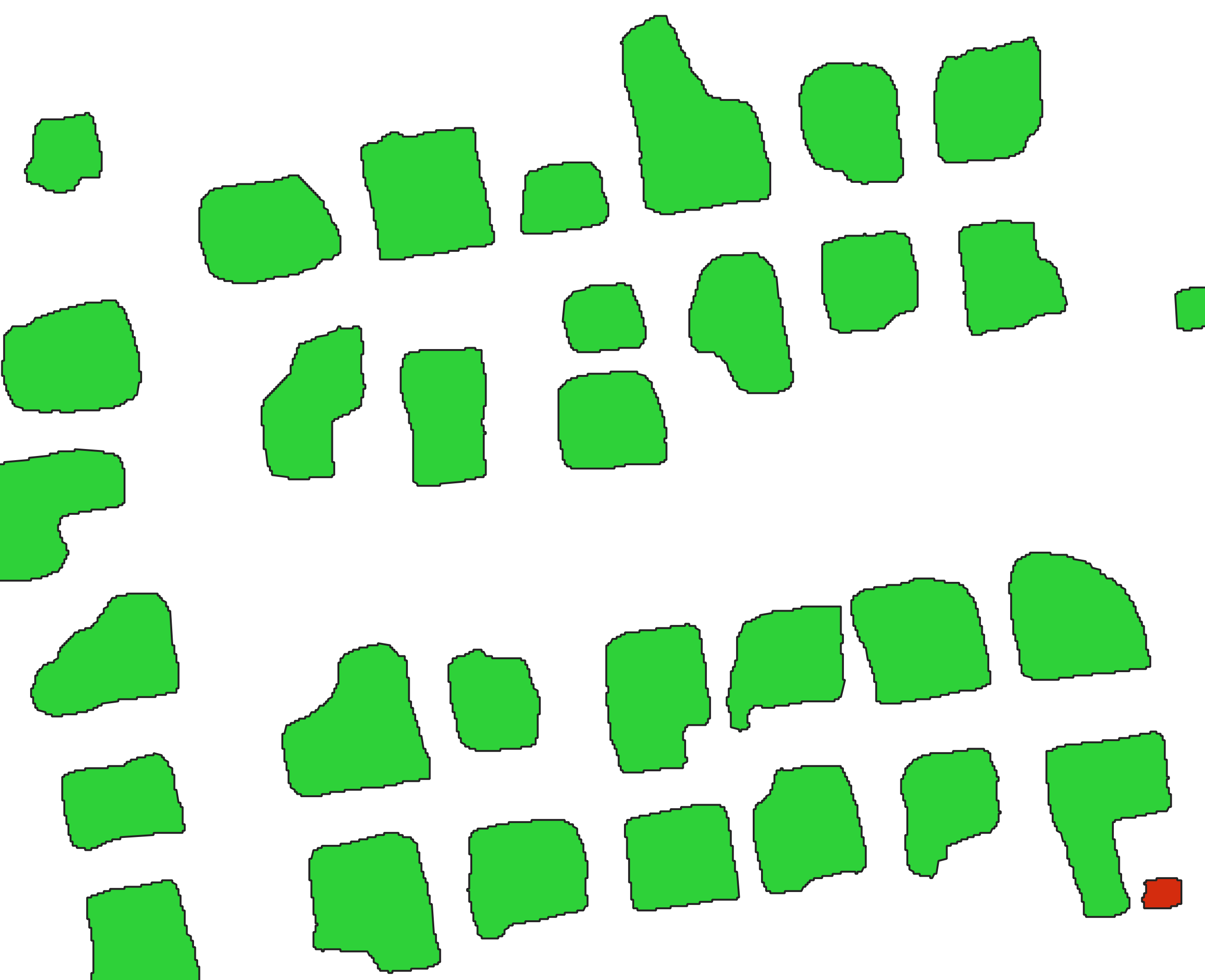} \\
 \hline & & & &\\
 \rotatebox{90}{\hspace{3.5 mm} Image Tiling} &
 \includegraphics[height=0.15\linewidth, width=0.22\linewidth]{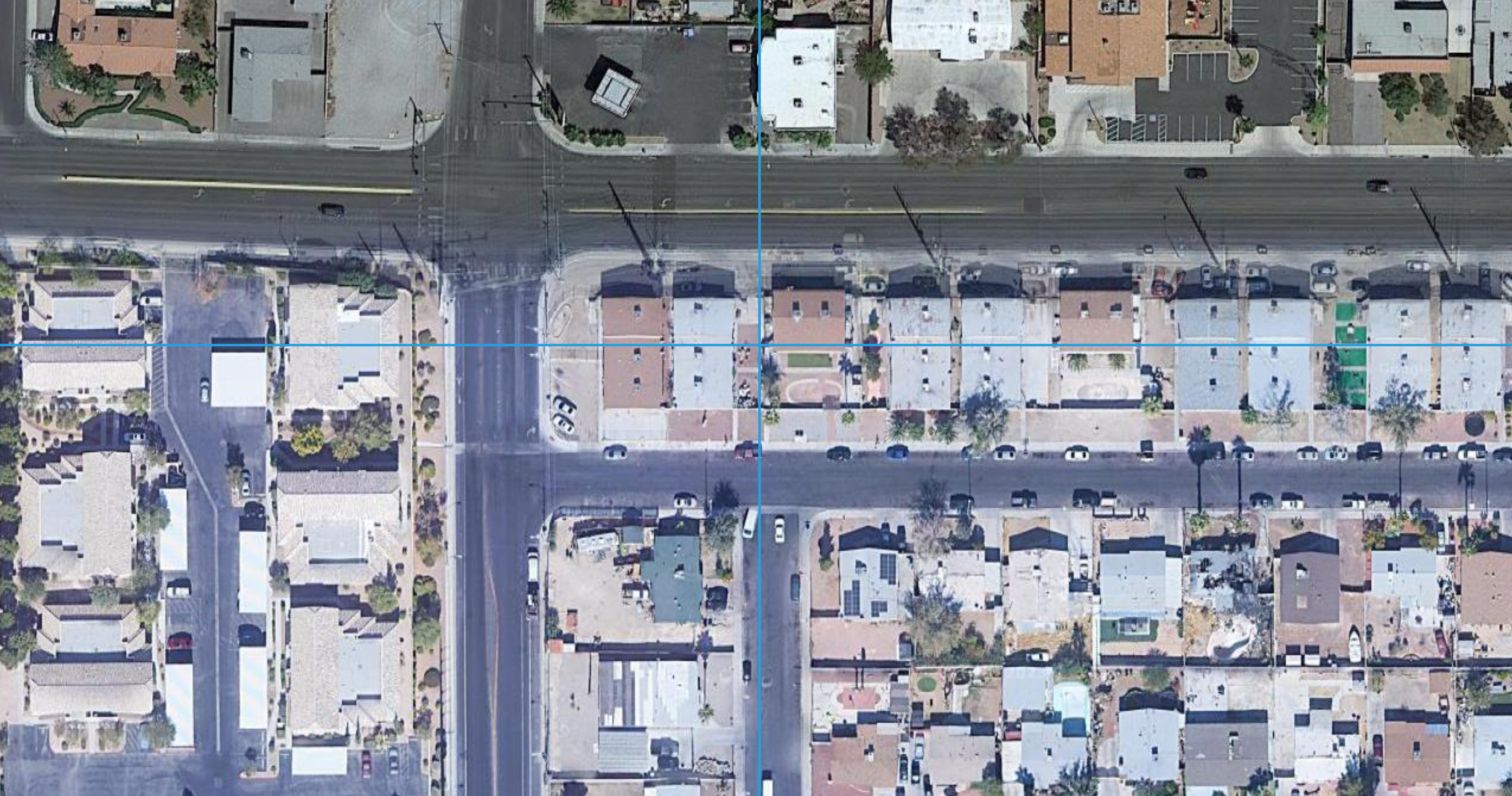} &
 \includegraphics[height=0.15\linewidth, width=0.22\linewidth]{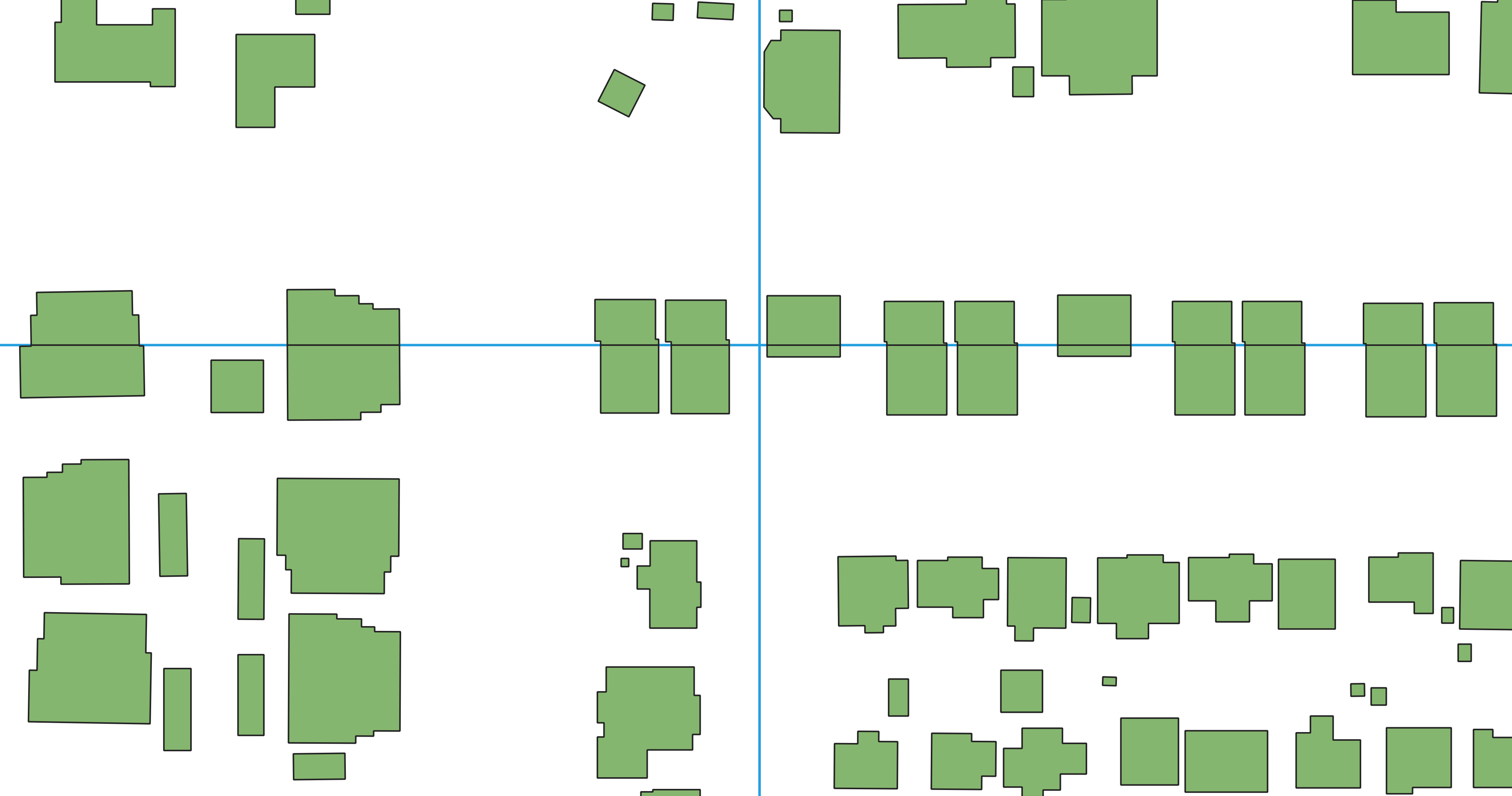} &
 \includegraphics[height=0.15\linewidth, width=0.22\linewidth]{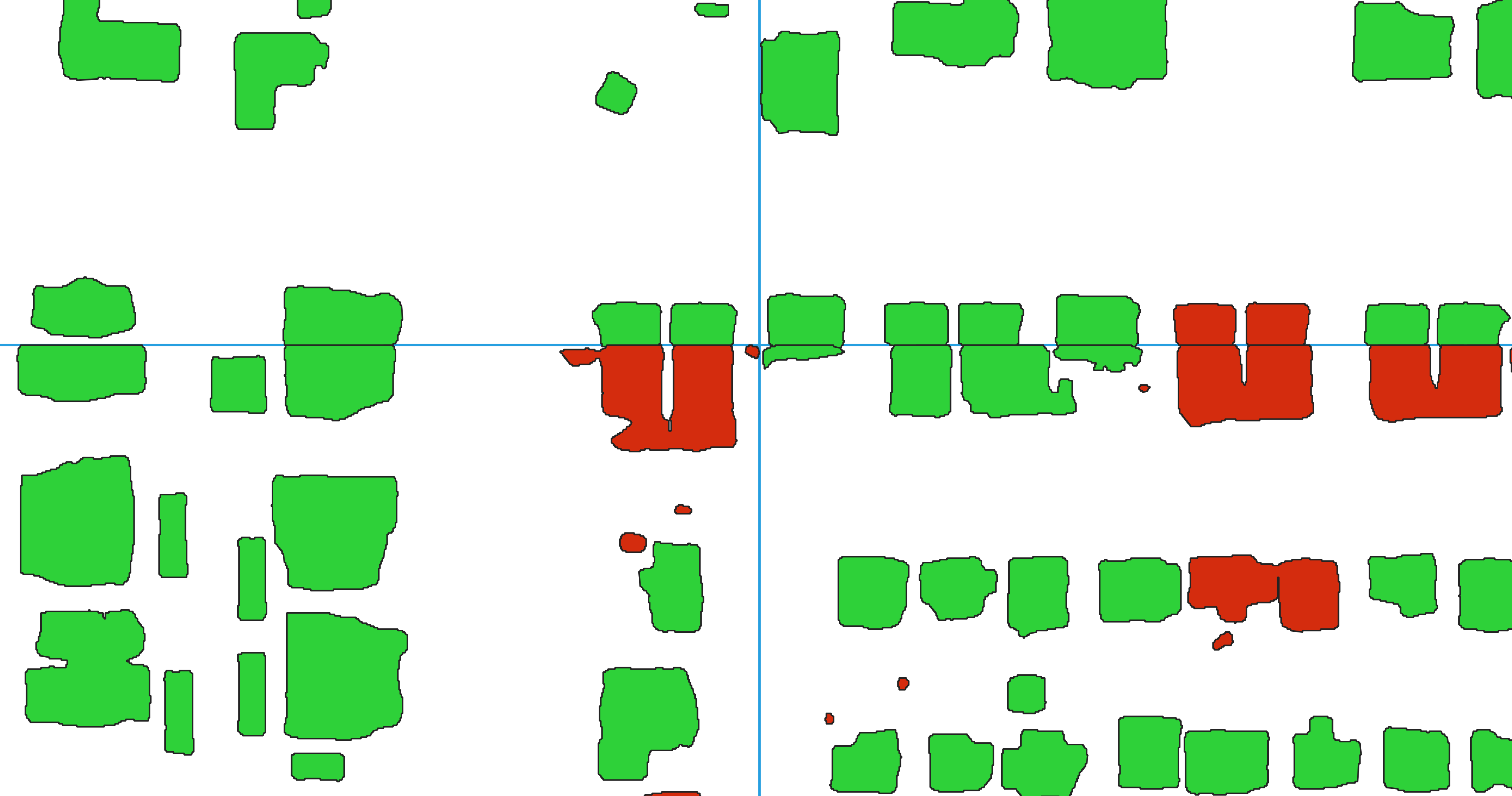} &
 \includegraphics[height=0.15\linewidth, width=0.22\linewidth]{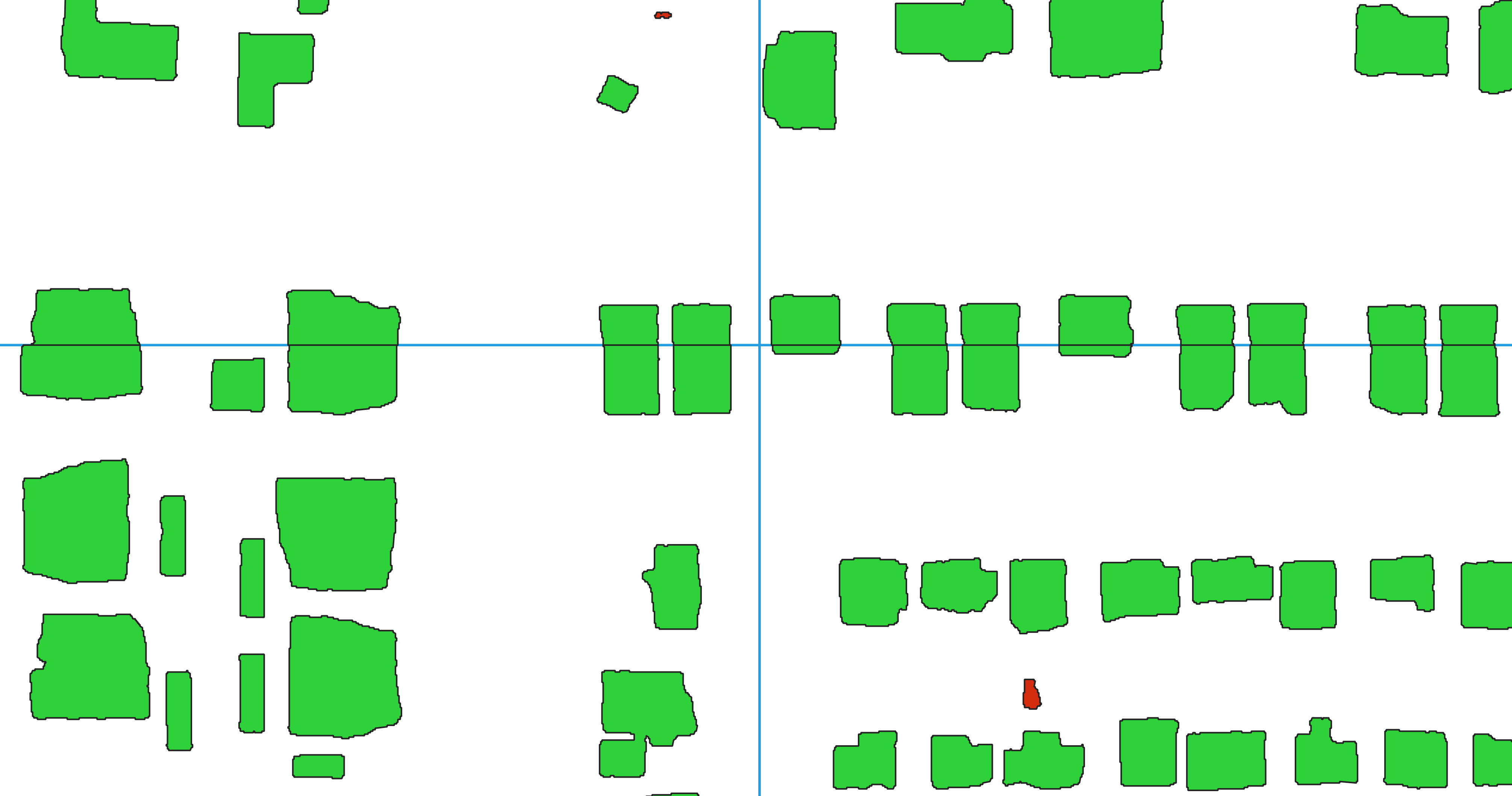}\\
 \hline\\
\end{tabular}
\caption{\textbf{Comparison of our proposed methodology with state-of-the-art (SOTA) methods}. In the first row, we show how SOTA model(s) fail to distinguish closely connected buildings from SpaceNet's Khartoum region, while our proposed design can effectively handle such situations. In the second row, we show the problems introduced at the image borders (blue grid lines show the division policy between consecutive tiles from the WHU Aerial dataset). The standard approach to train the model usually fails to perform well at such \textit{partially-cut} buildings. Using our pre-processing pipeline, deep models can easily capture such dynamics.}
\label{fig:IntroFig}
\end{figure*}

The availability of open-source building footprint datasets such as Toronto City \cite{wang2016torontocity}, INRIA \cite{maggiori2017can}, WHU \cite{ji2018fully}, ISPRS \cite{ISPRS}, Space-Net \cite{van2018spacenet}, xBD \cite{xBDdataset}, and AIcrowd \cite{mohanty2020deep} have allowed researchers to train deep learning models for building footprint extraction (BFE) that perform reasonably well. Indeed, the availability of these benchmark datasets has boosted research on automatic BFE from satellite imagery with numerous studies published in the last few years. The backbone of many of these studies, as will be discussed in Section \ref{sec:literature}, are deep segmentation models such as DeepLabV3+ \cite{chen2018encoder}. Despite these advances though, automated BFE from satellite imagery remains a challenging problem due to various reasons. For instance, buildings in different regions are characterized by high variability in shapes, sizes, and texture as well as occlusions due to surroundings or shadows. Another problem that arises with the use of semantic segmentation models is that they usually struggle to separate closely packed instances of building footprints in which the number of pixels separating two adjacent buildings is minimal. This problematic situation especially arises in developing countries. An instance of this scenario is depicted in the first row of  Fig. \ref{fig:IntroFig} indicating how state-of-the-art methods such as \cite{jiwani2021semantic} fails to distinguish closely packed buildings. Another challenge arises due to the tiling process of large tiff files -- the tiling step becomes necessary for training of deep learning models. In essence, splitting a large file into tiles breaks the spatial relationships at the boundaries, thus impacting the building structures residing there. An instance of this scenario is depicted in the second row of Fig. \ref{fig:IntroFig} in which the benchmark state-of-the-art method fails to identify some buildings at the tile boundaries. 

In this paper, we try to overcome both problems mentioned above. For improved performance on densely packed buildings, we propose a new semantic segmentation network model based on the standard DeepLabV3+ architecture. The model comprises of a single encoder and two parallel-placed decoders to separately handle reconstruction of (a) the building footprint and (b) the building edge. This is motivated by our understanding that along with the building mask generation, the edge mask generation is also essential in extracting building footprints, and therefore both must be accorded due importance in the training process. While some studies (e.g., \cite{jiwani2021semantic}) have tried to incorporate building edge information into the training framework, they do so while sticking with a generic single encoder-decoder design. In contrast, considering the unique behaviors of the building segmentation and edge segmentation tasks, we propose for them to be handled by two distinct decoders working on the high-dimensional features output of a single encoder. Given this tuning fork-like arrangement, we coin and use the term Tuning-Fork Network (TFNet) for this novel design which is found to perform well not only for scenarios with closely-packed buildings, but also for widely spaced building construction patterns. In addition to the improved TFNet design, we also propose a novel methodology that incorporates neighborhood pixels of the tile boundaries during training. The aim of this pre-processing pipeline, termed \textbf{Ne}ighborhood \textbf{P}ixel \textbf{AGG}regation (NePAGG), is to avoid the loss of spatial connectivity at the image borders. Armed with these two proposed novelties, we test our design on the WHU and SpaceNet2 benchmark datasets as well as a dataset we create from a region in the city of Lahore, Pakistan. The proposed model is shown to significantly outperform existing benchmark methods for all datasets.

In summary, the overall contributions of this work are as follows:

\begin{enumerate}
\item{We propose a novel pre-processing pipeline, named NePAGG, which incorporates spatial neighbourhood information during training.}
\item{We propose a novel architecture for BFE, namely TFNet that uses building edge information in addition to the standard footprint. We find that this results in improved detection performance.}
\item{We  create a new challenging dataset for footprint extraction for the city of Lahore in Pakistan, which is available publicly\footnote{\url{https://github.com/Muhammad-Ahmad-Waseem/TF-Net/tree/main}}.} 
\item{The proposed methodology is shown to achieve excellent performance on the standard SpaceNet \cite{SNweb}. and the WHU \cite{WHUweb} datasets, as well as the dataset we create for Lahore, Pakistan.}
\end{enumerate}

The remainder of this paper is organized as follows. \mbox{Section \ref{sec:literature}} provides a review of related work appearing in recent literature. Section \ref{sec:methodology} provides details of the proposed TFNet model and the NePAGG pre-processing methodology. This is followed by details of the experiments utilized for performance evaluation in Section \ref{sec:experiments}, while Section \ref{sec:results} presents the evaluation results. Section \ref{sec:conclusion} concludes the paper.

\begin{figure*}[h!]
\centering
 \includegraphics[width=0.9\linewidth]{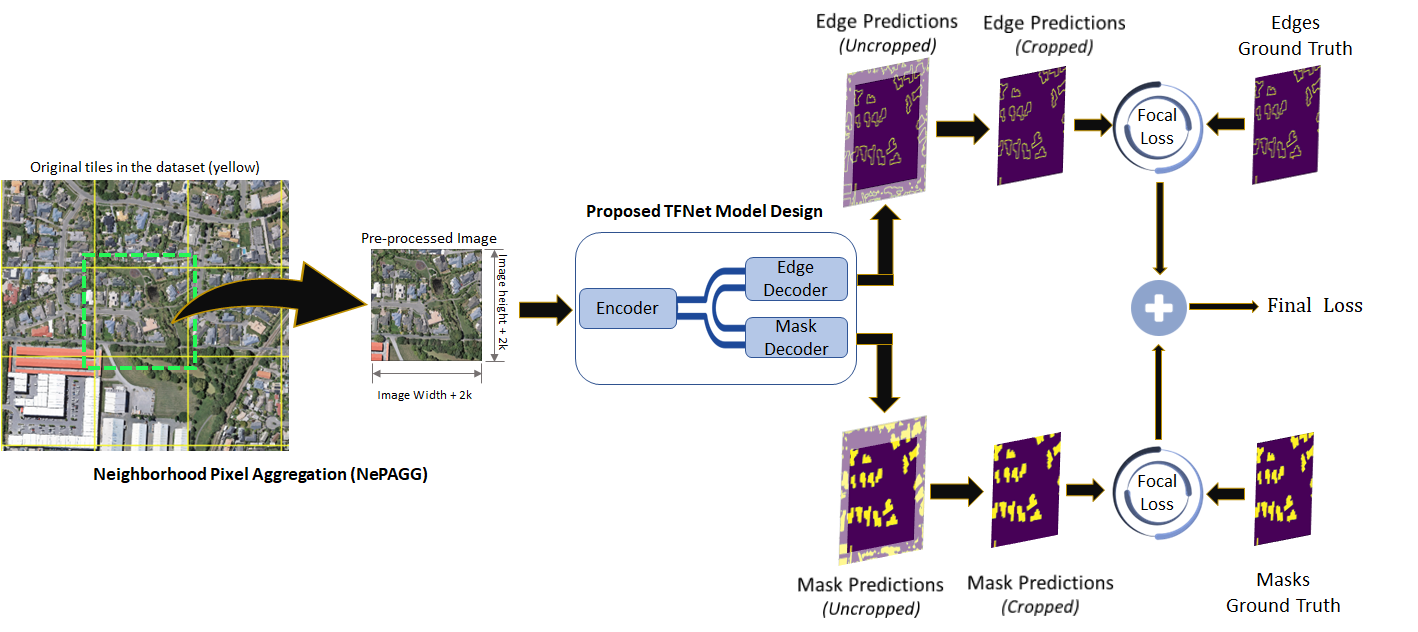}
\caption{\textbf{Proposed Methodology}. We first pre-process the images to include neighborhood pixels in model training. The pre-processed image is used during training, while loss is computed using the cropped part (original pixel locations \textit{before} pre-processing). The final loss is the sum of both focal losses of edges and masks.}
\label{fig:pipeline}
\end{figure*}

\section{Related Work}\label{sec:literature}
As mentioned earlier, the BFE problem  has been well studied in the literature. Initially, studies attempting BFE from satellite imagery relied on classical computer vision methods \cite{article}, \cite{6509456}, \cite{7524010}, \cite{gao2018building}. Although these classical approaches needed less training data, they struggled to deliver effective results since they were unable to handle complicated scenarios, e,g., varying architectural designs and similarity of building features with other features like roads. These classical approaches were followed by studies that use the generalization ability of deep learning models such as Convolutional Neural Networks (CNNs) to develop effective deep computer vision approaches for BFE \cite{wen2019automatic},  \cite{zhao2018building}, \cite{li2021building}, \cite{jiwani2021semantic}, \cite{8453267}, \cite{shi2018building}, \cite{8554076}, \cite{rs11070830}, \cite{8428417}. In recent years, semantic segmentation models have shown immense potential in providing effective solutions to automated BFE. Indeed, the winning solutions for most of the open challenges (such as SpaceNet2 and SpaceNet7) have been based on semantic segmentation models. Many authors have used baseline segmentation models, such as UNet \cite{ronneberger2015u}, SegNet \cite{badrinarayanan2017segnet}, FC-DenseNet \cite{jegou2017one}, DeepLabV3+ \cite{chen2018encoder}, and HRNet \cite{sun2019high} to extract building footprints from high-resolution satellite imagery. In addition to these, modified architectures for the encoder, decoder, or bottleneck modules have also been proposed that address the problem of semantic segmentation specifically for BFE \cite{zhu2020map}, \cite{chen2022asf}, \cite{liu2019net}. An important example of the utilization of segmentation models is the Google's Open Buildings dataset\cite{OpenBuildingsGoogle}. The dataset, now made publicly available for Africa, South Asia, South-East Asia, Latin America and the Caribbean, is very rich indeed. The first version of Google's dataset was generated using a simple simple U-net architecture \cite{sirko2021continental}. However, details of the model used to generate the latest version of the dataset, released very recently, are not publicly available. Similarly, the corresponding satellite imagery against this dataset is also not available which makes it difficult to train new models on these footprints because of the satellite imagery offsets \cite{offsetissue1}. This limits the applicability of the publicly available information for applications such as urban change detection and disaster damage assessment.\\

Although simple semantic segmentation models have shown excellent performance in extracting building footprints, many experiments show that these predicted masks are in need of improvement when the buildings are packed close together such that the building boundary is not clearly visible. To solve this problem and improve the performance in such locations, many authors have proposed to use building edge information during training. For example, \cite{yuan2017learning} proposed a signed distance transform (SDT) that computes for each pixel the distance from the building boundary. The SDT is then encoded into a finite number of classes followed by treating it as a multi-class segmentation problem. On the other hand, the work in \cite{jiwani2021semantic} proposed a weighted edge mask based on building boundary that assigns a higher weight to the loss generated from pixels residing at the building boundary. A related work is \cite{li2021building} that achieves improved performance using attraction field maps that are generated based on building boundaries during the training process. These methods show the importance of incorporating building edge information during the training of deep learning models for BFE. To the best of our knowledge though, the TFNet design consisting of a single encoder followed by two separate decoders for building and edge segmentation has not been explored in the literature before.

\section{Proposed Methodology}\label{sec:methodology}
Deep-learning-based methods for BFE from satellite imagery involve the splitting of large raster and vector files into small tiles, followed by training of the deep learning model on those small tiles for predicting footprints. Most proposals in the literature follow a standardized method of splitting the larger raster into non-overlapping tiles with each tile fed into the deep learning model independently of the others. A problem with this approach is that it fails to account for the spatial relationships between the edges of adjacent tiles. It is to address this issue that we propose a novel pipeline called NePAGG to incorporate neighborhood information of each tiled image. Secondly, to improve BFE especially for densely packed buildings, we propose TFNET, a novel single encoder dual-decoder network that incorporates the building edges' information during training. An overview of the proposed methodology is shown in Fig. \ref{fig:pipeline} with the two novel components described in the subsections below.

\subsection{NePAGG Pipeline}
NePAGG, as the name suggests, is the proposed technique that attempts to incorporate neighborhood information of each tile while training the deep learning model. To illustrate the loss of spatial information in traditional tiling, we refer the reader to the example shown in the bottom row of Fig. \ref{fig:IntroFig} in which a bigger raster is split into four non-overlapping tiles. As can be seen, there are several buildings present at the tile boundaries. Since training and prediction on each tile is carried out independently of others, the conventional method of having non-overlapping tiles results in the subsequent deep network to see partial building footprints at the boundaries, thus impairing its learning and prediction capability on the partial view of these buildings. It is to overcome this limitation that we propose the simple yet effective NePAGG methodology. As seen in Fig. \ref{fig:pipeline}, NePAGG utilizes overlapping tiles obtained by including neighborhood pixels at the boundaries. In particular, if the size of each tile with conventional non-overlapping tiling was $W \times H$ pixels, NePAGG picks up an extra $k$ pixels in the neighborhood of each boundary to obtain an augmented tile of size $(W + 2k) \times (H + 2k)$ pixels. This augmented tile is then used as input to the deep learning model that outputs segmentation masks of the same size. The segmentation masks are then cropped to the original size of $W \times H$ pixels before computing the training loss. The aforementioned pipeline ensures that some neighborhood information is available for each image tile at the boundaries resulting in better prediction. 

The choice of an appropriate $k$ in NePAGG depends on the resolution of the satellite imagery. A very large value of $k$ will produce large images as input to the deep learning model resulting in inefficient utilization of memory resources. On the other hand, a very small value of $k$ may not be enough to reap the benefits of neighborhood pixel aggregation. An appropriate tradeoff is struck when $k$ is just enough for complete building footprints to be included in the augmented tile.

\begin{figure*}[t]
\centering
 \includegraphics[width=0.9\linewidth]{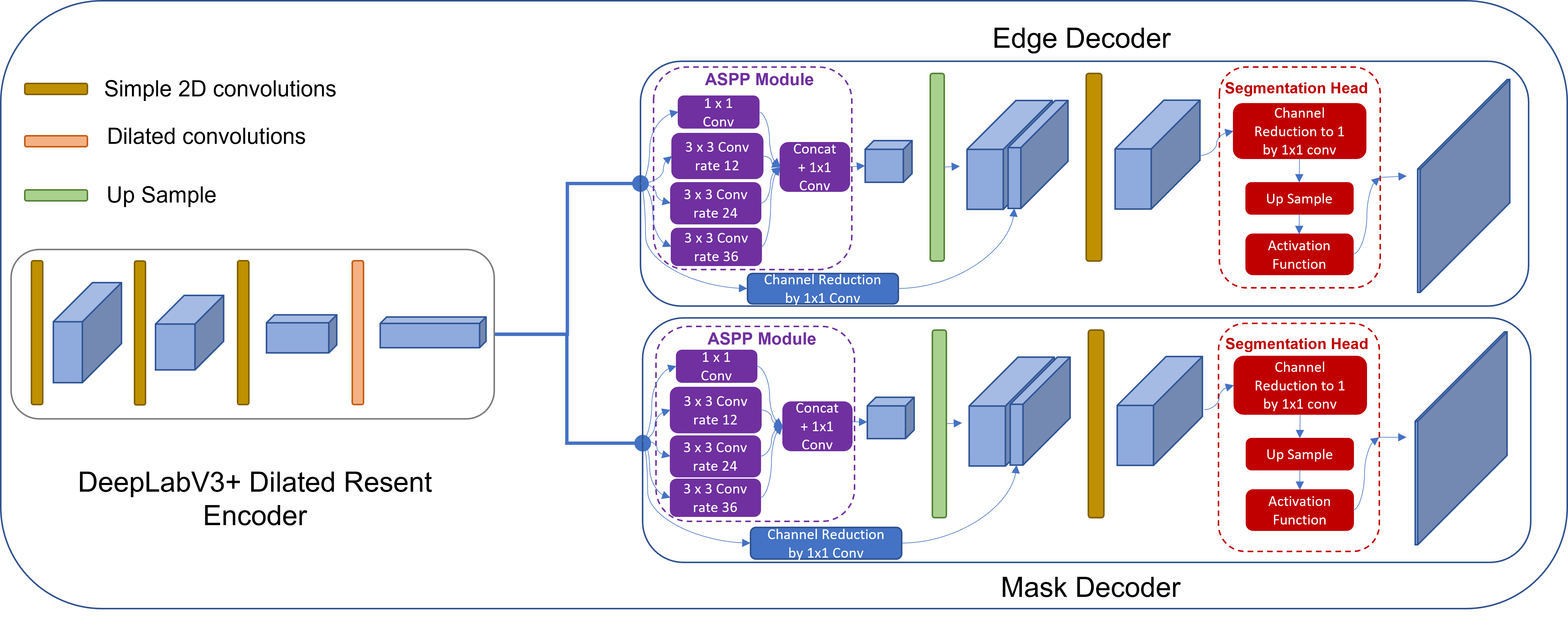}
\caption{\textbf{TFNet Detailed Architecture}. The encoder is a dilated ResNet encoder, while each decoder has a standard architecture with ASPP modules, as proposed in the DeepLabV3+ model \cite{chen2018encoder}.}
\label{fig:TFNet}
\end{figure*}

\subsection{TFNet Model Design}
As discussed earlier, the standard single encoder-decoder designs of semantic segmentation models usually fail to perform well when the distance between consecutive instances of buildings is small. In such cases, there is a need to focus the model's predictions at the building edge. As mentioned in Section \ref{sec:literature}, existing literature attempts to incorporate building edge information in the training pipeline through a variety of methods. As opposed to existing work, the proposed TFNet model involves a novel architecture composed of a single encoder and two decoders; one for generating the building masks and the other to separately deal with generating the building edge masks. 

The basic function of a decoder in a segmentation network is to learn the projection of dense features learned by the encoder for generation of segmentation masks. Introducing two separate decoders allows the model to learn this projection separately for building masks and the building edges. Moreover, since we train the model in an end-to-end fashion, both decoders also aid each other in the learning process through a back-propagation feedback loop, which also results in more refined feature space learning from encoders. It is due to these reasons that we expect our model to efficiently handle footprint extraction in high-density areas. Indeed, this is verified by the evaluation results presented in subsequent sections. \\

Considering the ability of atrous (or dilated) convolutions to better capture local and global relationships by providing an increased receptive field, TFNet uses a dilated ResNet network (proposed in the DeepLabv3+ segmentation model \cite{chen2018encoder}) as the backbone architecture for extracting semantic features from images. Similarly, for the decoder, an atrous spatial pyramid pooling module is used so as to allow several parallel atrous convolutions with different rates. This helps the model in preserving coarse or high-level semantic features with minimal computation power \cite{chen2018encoder}. Both of the decoder networks share exactly the same architecture with only difference being the corresponding ground truth that is used for training. The detailed architecture is shown in Fig. \ref{fig:TFNet}.

\subsection{Training Strategy}
As shown in Fig. \ref{fig:pipeline}, We propose an end-to-end pipeline for training of the deep learning model. The augmented tiles of size  $(W + 2k) \times (H + 2k)$ pixels are passed to the deep learning model. Since the semantic segmentation models are based only on convolutional layers, each decoder produces an output of exactly the same dimension as the input image. The images at the output of the two decoders are cropped to obtain the predicted segmentation masks of buildings and edges, each of size  $W  \times H$ pixels. Each of the predicted segmentation masks is compared to its corresponding ground truth to compute a focal loss. The sum of these two losses is then used to train the network in an end-to-end fashion.

\section{Experimentation}\label{sec:experiments}
\subsection{Datasets}
For performance comparisons, we utilize two benchmark datesets. In addition, we also provide a new dataset for Pakistan which is primarily meant to gauge performance in high-density urban areas. Some details about these datasets follow.

\subsubsection{SpaceNet2}
SpaceNet \cite{van2018spacenet} currently has three challenges that cover building footprints: SN1, SN2, and SN7. For this study, we use the most commonly used SN2 dataset which provides building footprints for four Areas of Interest (AOI) around the globe: Las Vegas, Paris, Shanghai, and Khartoum. Each training tile in the dataset covers 200 m $\times$ 200 m on the ground, and for each training tile, a 30 cm resolution geo-referenced tiff file is provided that has a size of $650 \times 650$ pixels. In addition, a geojson file containing ground truth markings for that region are also provided. The datasets are available online at \cite{SNweb}.\\


\subsubsection{WHU Aerial Images}
This dataset comes from the New Zealand Land Information Services who have generated the dataset by manually editing Christchurch's building vector data with about 22,000 independent buildings. The original ground resolution of the images is 0.075 m, but most of the aerial images are downsampled to a ground resolution of 0.3 m and subsequently cropped into 8,180 tiles of size $512 \times 512$ pixels each. The shapefile is also rasterized. The ready-to-use samples are divided into three parts:\\
\begin{itemize}
\item{A training set (130,500 buildings).}
\item{A validation set (14,500 buildings).} 
\item{A test set (42,000 buildings), which is further divided into Test1 and Test2 subsets}.
\end{itemize}
Accompanying the images are manually edited shapefile corresponding to the whole area. Since the dataset does not provide geo-referenced rasters, we used the order of cropping and area split shapefiles (available on the website) to recover the Geo-Tiff images and geojson masks against each image. The data can be downloaded from the Linz official website \cite{WHUweb}. \\


\subsubsection{Lahore DHA Dataset}
\begin{figure} [h]
    \centering
    \includegraphics[width=0.8\linewidth]{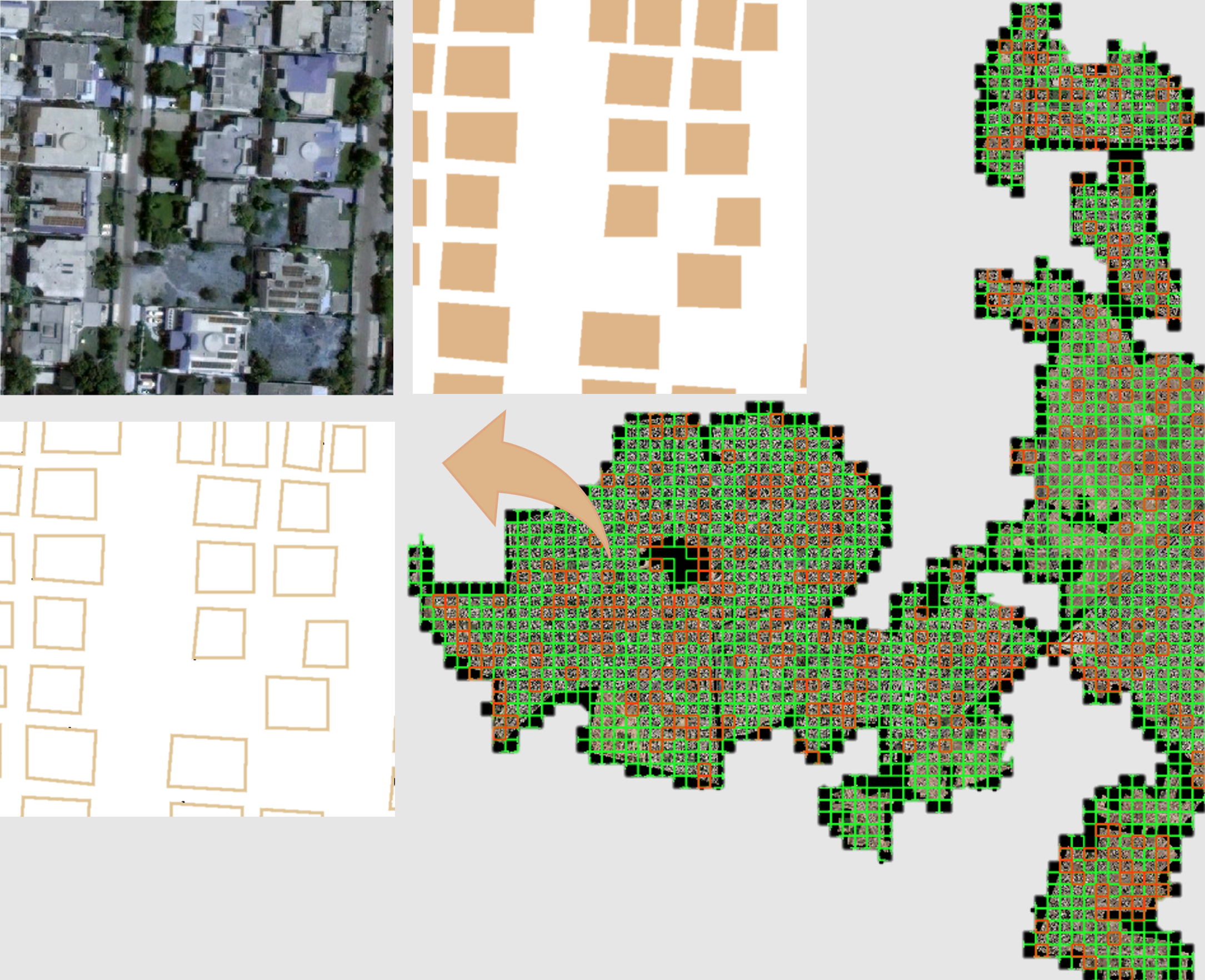}
    \caption{\textbf{Lahore DHA dataset details}. The green tiles represent the ones used for training while the red ones indicate those used for testing. A sample image from one of the tiles and the corresponding ground truth mask (for both the building footprint as well as the edges) are also shown.}
    \label{fig:citydata}
\end{figure}

We create a new dataset of building footprints for Pakistan that captures the complex and unique dynamics of the area. The dataset, illustrated in Fig. \ref{fig:citydata}, consists of 30 cm per pixel resolution images for a portion of the Defense Housing Authority (DHA) area in Lahore, Pakistan. The dataset comes with corresponding annotations of 25,631 building footprints covering an approximate area of 31 square km. The said area is divided into tiles of size $512 \times 512$ pixels, with the tiles then randomly split into training and testing portions with the following proportion:
\begin{itemize}
    \item \textbf{Train Tiles}: 1675, roughly 84\% of the total.
    \item \textbf{Test  Tiles}: 315,  roughly 16\% of the total.
\end{itemize}

\subsection{Training Parameters}
Our experiments are conducted within a Pytorch framework on an NVIDIA RTX 3090 GPU with 24 GB of memory. For the model training, remote sensing images are tiled, processed through the proposed NePAGG pipeline explained in \mbox{Section \ref{sec:methodology}}, before being fed into the training pipeline. The tile sizes used for training the models for each of the above datasets are described in the text below. 

\subsubsection{SpaceNet2} As described earlier, the SpaceNet dataset provides images of size $650 \times 650$ pixels. To make these divisible by 16, we perform zero padding at the edges to convert them to images of size $656 \times 656$ pixels. For benchmarking performance gain due to NePAGG, we train the TFNET model without it. For the purpose, the zero-padded image tiles of size $656 \times 656$ are used. On the other hand, when utilizing NePAGG, we fix $k = 83$ so that the final image tile used for training is of size $816 \times 816$ pixels. 

\subsubsection{WHU} The WHU dataset provides images of size ${512 \times 512}$ pixels.  For training with NePAGG, we set $k=64$ for this dataset to obtain a final image of size $640 \times 640$ pixels.  

\subsubsection{Lahore DHA} The dataset for Lahore DHA has exactly the same specifications as the WHU dataset, so we chose the same image sizes, i.e., $640 \times 640$ with NePAGG, and $512 \times 512$ without.

In addition to the segmentation masks for buildings in each dataset, we automatically generate building edge masks by selecting a few pixels from the boundaries of the building polygons. A sample depiction of this for the Lahore DHA dataset is shown in Fig. \ref{fig:citydata}. All models are trained for 150 epochs on each dataset, and the optimizer is stochastic gradient descent (SGD) with a learning rate of 0.0001. The training batch size of all models is set to 8. As mentioned earlier, the Focal Loss is used as a loss function for our model.


\renewcommand{\arraystretch}{1.2}
\begin{table*}
    \centering
    \caption{\textbf{Quantitaive Comparison on SpaceNet2 Dataset.} Accuracy of different methods on different cities of SN2. We train a single model for all cities. The results correspond to the whole train-split provided by SpaceNet2, as ground-truth marking for the test set is not available.} 
    \begin{tabular}{*{6}{|wc{2cm}}|}
     \hline    
       \multirow{2}{*}{Metric} & \multirow{2}{*}{Method} & \multicolumn{4}{|c|}{AOI} \\ \cline{3-6}
       & & Vegas & Paris & Shanghai & Khartoum \\ \hline
       \multirow{4}{*}{{Precision}} & Deeplabv3+\cite{chen2018encoder} & 89.37 & 64.40 & 60.00 & 55.22\\\cline{2-6}
                                 & rgb footprint \cite{jiwani2021semantic} & 90.89 & 75.93 & 66.99 & 66.12 \\\cline{2-6}
                                 & TFNet w/o NePAGG & 95.59 & 83.46 & 79.28 & 77.04 \\\cline{2-6}
                                 & TFNet with NePAGG & \textbf{97.82} & \textbf{91.56} & \textbf{88.32} & \textbf{87.14}\\\hline
        \multicolumn{6}{|c|}{} \\ \hline
      \multirow{4}{*}{{Recall}} & Deeplabv3+\cite{chen2018encoder} & 77.23 & 48.22 & 35.35 & 35.81\\\cline{2-6}
                                 & rgb footprint \cite{jiwani2021semantic} & 86.83 & 66.67 & 49.57 & 48.52 \\\cline{2-6}
                                 & TFNet w/o NePAGG & 88.19 & 72.08 & 61.18 & 61.77 \\\cline{2-6}
                                 & TFNet with NePAGG & \textbf{92.97} & \textbf{83.08} & \textbf{74.40} & \textbf{75.43}\\\hline
        \multicolumn{6}{|c|}{} \\ \hline
        \multirow{4}{*}{{F1 Score}} & Deeplabv3+\cite{chen2018encoder} & 82.85 & 55.15 & 44.49 & 43.44\\\cline{2-6}
                                 & rgb footprint \cite{jiwani2021semantic} & 88.81 & 71.00 & 56.97 & 55.97 \\\cline{2-6}
                                 & TFNet w/o NePAGG & 91.74 & 77.35 & 69.06 & 68.57 \\\cline{2-6}
                                 & TFNet with NePAGG & \textbf{95.33} & \textbf{87.11} & \textbf{80.76} & \textbf{80.86}\\\hline
    \end{tabular}
    \label{tab:SN2}
\end{table*}

\begin{table*}
    \centering
    \caption{\textbf{Quantitaive Comparison on WHU Dataset.} Accuracy of different methods on different splits of WHU. We train the model using the \emph{Train} split and validate the results using the \emph{Val} split of the dataset. No image from the \emph{Test} or the \emph{Test2} split was used during the training of any of the provided models.}
    \begin{tabular}{*{6}{|wc{2cm}}|}
     \hline    
       \multirow{2}{*}{Metric} & \multirow{2}{*}{Method} & \multicolumn{4}{|c|}{Split} \\ \cline{3-6}
       & & Train & Val & Test & Test2 \\ \hline
       \multirow{4}{*}{{Precision}} & Deeplabv3+\cite{chen2018encoder} & 92.65 & 93.39 & 92.84 & 93.31\\\cline{2-6}
                                 & rgb footprint \cite{jiwani2021semantic} & 94.22 & 91.89 & 89.12 & 92.27 \\\cline{2-6}
                                 & TFNet w/o NePAGG & 95.94 & 95.05 & 93.54 & 95.23 \\\cline{2-6}
                                 & TFNet with NePAGG & \textbf{98.57} & \textbf{96.00} & \textbf{94.43} & \textbf{96.09}\\\hline
        \multicolumn{6}{|c|}{} \\ \hline
      \multirow{4}{*}{{Recall}} & Deeplabv3+\cite{chen2018encoder} & 72.01 & 75.63 & 70.00 & 72.75\\\cline{2-6}
                                 & rgb footprint \cite{jiwani2021semantic} & 83.91 & 85.34 & 80.79 & 84.18 \\\cline{2-6}
                                 & TFNet w/o NePAGG & 84.34 & 85.25 & 80.36 & 84.26 \\\cline{2-6}
                                 & TFNet with NePAGG & \textbf{91.27} & \textbf{89.74} & \textbf{85.40} & \textbf{89.31}\\\hline
        \multicolumn{6}{|c|}{} \\ \hline
        \multirow{4}{*}{{F1 Score}} & Deeplabv3+\cite{chen2018encoder} & 81.04 & 83.57 & 79.82 & 81.76\\\cline{2-6}
                                 & rgb footprint \cite{jiwani2021semantic} & 88.77 & 88.49 & 84.75 & 88.04 \\\cline{2-6}
                                 & TFNet w/o NePAGG & 89.77 & 89.88 & 86.45 & 89.40 \\\cline{2-6}
                                 & TFNet with NePAGG & \textbf{94.78} & \textbf{92.77} & \textbf{89.69} & \textbf{92.58}\\\hline
    \end{tabular}
    \label{tab:WHU}
\end{table*}

\begin{table*}
    \centering
    \caption{\textbf{Quantitaive Comparison on Lahore DHA Dataset.} Accuracy of different methods on different splits of the Lahore DHA dataset. We train the model using the \emph{Train} split. No image from the \emph{Test} split was used during training of any of the provided models.}
    \begin{tabular}{*{4}{|wc{2.5cm}}|}
     \hline    
       \multirow{2}{*}{Metric} & \multirow{2}{*}{Method} & \multicolumn{2}{|c|}{Split} \\ \cline{3-4}
       & & Train & Test \\ \hline
       \multirow{4}{*}{{Precision}} & Deeplabv3+\cite{chen2018encoder} & 34.81 & 35.05\\\cline{2-4}
                                 & rgb footprint \cite{jiwani2021semantic} & 63.39 & 46.51 \\\cline{2-4}
                                 & TFNet w/o NePAGG & 99.62 & 71.98  \\\cline{2-4}
                                 & TFNet with NePAGG & \textbf{99.73} & \textbf{78.70}\\\hline
        \multicolumn{4}{|c|}{} \\ \hline
      \multirow{4}{*}{{Recall}} & Deeplabv3+\cite{chen2018encoder} & 21.45 & 20.56\\\cline{2-4}
                                 & rgb footprint \cite{jiwani2021semantic} & 31.90 & 25.25 \\\cline{2-4}
                                 & TFNet w/o NePAGG & 97.68 & 68.92  \\\cline{2-4}
                                 & TFNet with NePAGG & \textbf{98.19} & \textbf{76.45}\\\hline
        \multicolumn{4}{|c|}{} \\ \hline
        \multirow{4}{*}{{F1 Score}} & Deeplabv3+\cite{chen2018encoder} & 26.55 & 25.92\\\cline{2-4}
                                 & rgb footprint \cite{jiwani2021semantic} & 42.44 & 32.73 \\\cline{2-4}
                                 & TFNet w/o NePAGG & 98.64 & 70.41\\\cline{2-4}
                                 & TFNet with NePAGG & \textbf{98.95} & \textbf{77.56}\\\hline
    \end{tabular}
    \label{tab:CITY}
\end{table*}
\renewcommand{\arraystretch}{1}

\subsection{Evaluation Metrics}
Most existing methods for BFE using semantic segmentation models provide pixel-based accuracy for performance evaluation. Although it is a good metric for evaluation of pixel-based prediction tasks, it does not provide adequate classification accuracy information at the building level. For this purpose, we resort to the standard metrics used in SpaceNet challenges that are based on the \emph{number of polygons} identified correctly/incorrectly.\\ 

To evaluate the performance of the models, we use the standard F1 Score computed on the basis of the \textit{number of polygons}, instead of the image pixels. For each of the predicted image, we create polygons by creating a geometry over the \textit{connected regions} of pixels. This can be easily done using the scikit image processing tools and the rasterio geospatial libraries available in python. Let $\mathcal{P} = \{P_1,P_2, ..., P_M\}$ be the set of predicted polygons obtained from these images and $\mathcal{L} = \{L_1,L_2, ..., L_N\}$ be the set of corresponding ground truth polygons. Each one of the predicted polygon is labelled as True Positive or False Positive based on the Intersection over Union (IoU) scores between the predicted and the ground truth polygons. Details of this evaluation method, as used in SpaceNet's challenges, are provided in Algorithm \ref{alg:SN}.\\ 

\begin{algorithm}
\caption{SpaceNet Evaluation Method}
\begin{algorithmic}[1]
\State $i \gets 1$
\State $\mathcal{P} \gets \{P_1,P_2, ..., P_M\}$
\State $\mathcal{L} \gets \{L_1,L_2, ..., L_N\}$ 

\While{$i \leq M$} 

\State $S_{i} \gets \smash{\displaystyle\max_{j\in \{1,\ldots,N\}}} \{\text{IoU}(P_i, L_j)\}$\\
\State $k \gets \smash{\displaystyle \arg \: \max_{j\in \{1,\ldots,N\}}} \{\text{IoU}(P_i, L_j)\}$\\
 
\If{$S_{i}\geq 0.5$}
    \State $P_i$ is True Positive
    \State Remove $L_k$ from $\mathcal{L}$
\Else
    \State $P_i$ is False Positive
\EndIf 
\State $i \gets i+1$
\EndWhile
\end{algorithmic}
\label{alg:SN}
\end{algorithm}


Once the predicted polygons are classified into true positives and false positives, all the remaining polygons in the list $\mathcal{L}$ represent missed detections / false negatives. The calculations for IoU, precision, recall and F1 Score are done using standard formulas as shown below:

\setlength{\arraycolsep}{0.0em}
\begin{eqnarray}
IoU(P_i, L_j) = \frac{A(P_i\;\cap\;L_j)}{A(P_i\;\cup\;L_j)}
\end{eqnarray}
\begin{eqnarray}
\rm{Precision} = \frac{\rm{TP}}{\rm{TP} + \rm{FP}}
\end{eqnarray}
\begin{eqnarray}
\rm{Recall} = \frac{\rm{TP}}{\rm{TP} + \rm{FN}}
\end{eqnarray}
\begin{eqnarray}
\rm{F1} = \frac{2\;x\;\rm{Precision}\;x\;\rm{Recall}}{\rm{Precision}\;+\;\rm{Recall}\;-\;\rm{Precision}\;x\;\rm{Recall}}
\end{eqnarray}
\setlength{\arraycolsep}{5pt}
\\
Here, $A(P)$ represents the area of a polygon $P$, while TP, FP, FN represent the total number of true positives, false positives, and false negatives, respectively.

\section{Results and Comparisons}\label{sec:results}
As discussed earlier, we have chosen three different datasets for comparison. For a fair analysis, we compare the performance of our method with the ones based on DeepLabV3+ architecture, which is considered as state-of-the-art semantic segmentation network. The first method that we compare with is a standard DeepLabV3+ model with Focal Loss \cite{chen2018encoder}. Similarly, we also show a comparison with RGB Footprints by Jiwani et. al \cite{jiwani2021semantic}, which tries to incorporate an edge-based loss with a standard DeepLabV3+ architecture. In addition to these, we also show the performance of our proposed TFNet design \textit{without} the NePAGG pre-processing pipeline. To make the comparisons consistent, all the above mentioned benchmark methods are also trained \textit{without} NePAGG, while our final proposed model is trained using both NePAGG and the TFNet design. In this way, we can quantitatively elaborate the importance of incorporating each of the proposed methodologies. The performance of each of the described methods is measured using evaluation metric and the training parameters described in Section \ref{sec:experiments}.\\


\subsection{Comparison on the SpaceNet2 Dataset}
For the SpaceNet2 dataset, we combine all the training images from four different AOIs into a single dataset and train the model using that dataset. An 80\% split of this combined dataset is used for training, while the remaining 20\% is used for validation and testing. Since SN2 does not provide ground-truth marking over the test sets, we only show the performance on the training portion of each AOI. The results for each AOI are shown in Table \ref{tab:SN2}. It can be seeen that while the proposed TFNet architecture alone is characterized by improved performance on all of the AOIs, the overall TFNet pipeline with NePAGG improves the performance to a new high. For each of the AOI, the proposed methodology obtains an F1 score of more than 80\% with an average F1 score of 86\% which is significantly superior in effectiveness over the benchmark methods. We also note that the performance gain of the proposed methodology over that of \cite{jiwani2021semantic} -- the closest performing benchmark in the literature -- is a lot higher for the cities of Shanghai and Khartoum as compared to Las Vegas and Paris. We posit that this is because of the buildings packed much closer together for Shangahi and Khartoum's dataset as compared to the other two cities.   

\subsection{Comparison on the WHU Dataset}
For the WHU dataset, we use the original \emph{Train} and \emph{Val} splits for training and validation of the model, respectively. The other two splits, i.e., \emph{Test} and \emph{Test2} were not used for training, and were just used to compute the accuracy scores for the unseen data split. The results on each of the splits are shown in Table \ref{tab:WHU}. Similar to the SpaceNet2 dataset, one can see from the table that while TFNet alone adds a performance boost on all of the splits, the proposed NePAGG plus TFNet design outperforms all of the provided methods on all of the different data splits. The results also indicate that our method not only performs well on the training data but also has improved performance over unseen areas.

\subsection{Comparison on the Lahore DHA Dataset}
For the Lahore DHA dataset we create ourselves, we only have two data splits, i.e., \emph{Train} and \emph{Test}. We use the train split for training and do not use any validation set for this dataset. The other split, i.e., the test split, is not used for training and is just used to check the score on unseen data split. The results are shown in Table \ref{tab:CITY}. The low score of the standard DeepLabv3plus \cite{chen2018encoder} model shows how challenging the dataset is for SOTA segmentation models. Compared to the benchmark methodologies in the literature, the proposed TFNet plus NePAGG methodology performs extremely well even in such complicated scenarios. We also note that the performance benefit because of the TFNet architecture is much higher compared to the additional gain obtained by NePAGG. 

\section{Conclusion}\label{sec:conclusion}
Considering the problem of BFE from satellite imagery, this paper proposes a novel methodology consisting of TFNet: a tuning-fork like encoder-decoder architecture coupled with NePAGG: a pixel aggregation pre-processing methodology. The TFNet architecture consists of a single encoder followed by two decoders (all based on the Deeplabv3 architecture) to separately detect detect building masks and building edges. This architecture specifically allows it to perform well on areas with densely connected buildings. On the other hand, NePAGG allows the model to incorporate spatial relationships of neighborhood images (at the tile boundaries) during the training phase. For performance evaluation, we utilize the polygon-based F1 scores and illustrate the  effectiveness of the proposed methodology on standard datasets such as SpaceNet2 and WHU Datasets, as well as a dataset we create ourselves for the city of Lahore in Pakistan. A potential direction for future research could be to explore the possibility of incorporating deep features of neighborhood images directly during training without altering the actual image size. Moreover, BFE for regions with unstructured / informal settlements still remains a challenging task. This remains a promising direction of future research. 




%
\bibliography{main.bib}{}

\begin{thebibliography}{10}
\providecommand{\url}[1]{#1}
\csname url@samestyle\endcsname
\providecommand{\newblock}{\relax}
\providecommand{\bibinfo}[2]{#2}
\providecommand{\BIBentrySTDinterwordspacing}{\spaceskip=0pt\relax}
\providecommand{\BIBentryALTinterwordstretchfactor}{4}
\providecommand{\BIBentryALTinterwordspacing}{\spaceskip=\fontdimen2\font plus
\BIBentryALTinterwordstretchfactor\fontdimen3\font minus
  \fontdimen4\font\relax}
\providecommand{\BIBforeignlanguage}[2]{{%
\expandafter\ifx\csname l@#1\endcsname\relax
\typeout{** WARNING: IEEEtran.bst: No hyphenation pattern has been}%
\typeout{** loaded for the language `#1'. Using the pattern for}%
\typeout{** the default language instead.}%
\else
\language=\csname l@#1\endcsname
\fi
#2}}
\providecommand{\BIBdecl}{\relax}
\BIBdecl

\bibitem{wang2016torontocity}
S.~Wang, M.~Bai, G.~Mattyus, H.~Chu, W.~Luo, B.~Yang, J.~Liang, J.~Cheverie,
  S.~Fidler, and R.~Urtasun, ``Torontocity: Seeing the world with a million
  eyes,'' \emph{arXiv preprint arXiv:1612.00423}, 2016.

\bibitem{maggiori2017can}
E.~Maggiori, Y.~Tarabalka, G.~Charpiat, and P.~Alliez, ``Can semantic labeling
  methods generalize to any city? the inria aerial image labeling benchmark,''
  in \emph{2017 IEEE International Geoscience and Remote Sensing Symposium
  (IGARSS)}.\hskip 1em plus 0.5em minus 0.4em\relax IEEE, 2017, pp. 3226--3229.

\bibitem{ji2018fully}
S.~Ji, S.~Wei, and M.~Lu, ``Fully convolutional networks for multisource
  building extraction from an open aerial and satellite imagery data set,''
  \emph{IEEE Transactions on Geoscience and Remote Sensing}, vol.~57, no.~1,
  pp. 574--586, 2018.

\bibitem{ISPRS}
``{ISPRS 2D Semantic Labeling Contest},'' accessed: May 27, 2022 [online].
  Available:
  "http://www2.isprs.org/commissions/comm3/wg4/semantic-labeling.html".

\bibitem{van2018spacenet}
A.~Van~Etten, D.~Lindenbaum, and T.~M. Bacastow, ``Spacenet: A remote sensing
  dataset and challenge series,'' \emph{arXiv preprint arXiv:1807.01232}, 2018.

\bibitem{xBDdataset}
\BIBentryALTinterwordspacing
R.~Gupta, R.~Hosfelt, S.~Sajeev, N.~Patel, B.~Goodman, J.~Doshi, E.~Heim,
  H.~Choset, and M.~Gaston, ``xbd: A dataset for assessing building damage from
  satellite imagery,'' 2019. [Online]. Available:
  \url{https://arxiv.org/abs/1911.09296}
\BIBentrySTDinterwordspacing

\bibitem{mohanty2020deep}
S.~P. Mohanty, J.~Czakon, K.~A. Kaczmarek, A.~Pyskir, P.~Tarasiewicz,
  S.~Kunwar, J.~Rohrbach, D.~Luo, M.~Prasad, S.~Fleer \emph{et~al.}, ``Deep
  learning for understanding satellite imagery: An experimental survey,''
  \emph{Frontiers in Artificial Intelligence}, vol.~3, 2020.

\bibitem{chen2018encoder}
L.-C. Chen, Y.~Zhu, G.~Papandreou, F.~Schroff, and H.~Adam, ``Encoder-decoder
  with atrous separable convolution for semantic image segmentation,'' in
  \emph{Proceedings of the European conference on computer vision (ECCV)},
  2018, pp. 801--818.

\bibitem{jiwani2021semantic}
A.~Jiwani, S.~Ganguly, C.~Ding, N.~Zhou, and D.~M. Chan, ``A semantic
  segmentation network for urban-scale building footprint extraction using rgb
  satellite imagery,'' \emph{arXiv preprint arXiv:2104.01263}, 2021.

\bibitem{SNweb}
``Spacenet.ai,'' Available at \url{https://spacenet.ai/datasets} (accessed May
  04, 2023).

\bibitem{WHUweb}
``Whu aerial data,'' Available at
  \url{https://data.linz.govt.nz/layer/51932-christchurch-post-earthquake-01m-urban-aerial-photos-24-february-2011/s}
  (accessed May 04, 2023).

\bibitem{article}
A.~Ok, C.~Senaras, and B.~Yuksel, ``Automated detection of arbitrarily shaped
  buildings in complex environments from monocular vhr optical satellite
  imagery,'' \emph{IEEE Transactions on Geoscience and Remote Sensing},
  vol.~51, pp. 1701--1717, 03 2013.

\bibitem{6509456}
X.~Huang, L.~Zhang, and T.~Zhu, ``Building change detection from multitemporal
  high-resolution remotely sensed images based on a morphological building
  index,'' \emph{IEEE Journal of Selected Topics in Applied Earth Observations
  and Remote Sensing}, vol.~7, no.~1, pp. 105--115, 2014.

\bibitem{7524010}
Q.~Zhang, X.~Huang, and G.~Zhang, ``A morphological building detection
  framework for high-resolution optical imagery over urban areas,'' \emph{IEEE
  Geoscience and Remote Sensing Letters}, vol.~13, no.~9, pp. 1388--1392, 2016.

\bibitem{gao2018building}
X.~Gao, M.~Wang, Y.~Yang, and G.~Li, ``Building extraction from rgb vhr images
  using shifted shadow algorithm,'' \emph{Ieee Access}, vol.~6, pp.
  22\,034--22\,045, 2018.

\bibitem{wen2019automatic}
Q.~Wen, K.~Jiang, W.~Wang, Q.~Liu, Q.~Guo, L.~Li, and P.~Wang, ``Automatic
  building extraction from google earth images under complex backgrounds based
  on deep instance segmentation network,'' \emph{Sensors}, vol.~19, no.~2, p.
  333, 2019.

\bibitem{zhao2018building}
K.~Zhao, J.~Kang, J.~Jung, and G.~Sohn, ``Building extraction from satellite
  images using mask r-cnn with building boundary regularization,'' in
  \emph{Proceedings of the IEEE conference on computer vision and pattern
  recognition workshops}, 2018, pp. 247--251.

\bibitem{li2021building}
Q.~Li, L.~Mou, Y.~Hua, Y.~Shi, and X.~X. Zhu, ``Building footprint generation
  through convolutional neural networks with attraction field representation,''
  \emph{IEEE Transactions on Geoscience and Remote Sensing}, vol.~60, pp.
  1--17, 2021.

\bibitem{8453267}
X.~Li, X.~Yao, and Y.~Fang, ``Building-a-nets: Robust building extraction from
  high-resolution remote sensing images with adversarial networks,'' \emph{IEEE
  Journal of Selected Topics in Applied Earth Observations and Remote Sensing},
  vol.~11, no.~10, pp. 3680--3687, 2018.

\bibitem{shi2018building}
Y.~Shi, Q.~Li, and X.~X. Zhu, ``Building footprint generation using improved
  generative adversarial networks,'' \emph{IEEE Geoscience and Remote Sensing
  Letters}, vol.~16, no.~4, pp. 603--607, 2018.

\bibitem{8554076}
J.~Hui, M.~Du, X.~Ye, Q.~Qin, and J.~Sui, ``Effective building extraction from
  high-resolution remote sensing images with multitask driven deep neural
  network,'' \emph{IEEE Geoscience and Remote Sensing Letters}, vol.~16, no.~5,
  pp. 786--790, 2019.

\bibitem{rs11070830}
\BIBentryALTinterwordspacing
P.~Liu, X.~Liu, M.~Liu, Q.~Shi, J.~Yang, X.~Xu, and Y.~Zhang, ``Building
  footprint extraction from high-resolution images via spatial residual
  inception convolutional neural network,'' \emph{Remote Sensing}, vol.~11,
  no.~7, 2019. [Online]. Available:
  \url{https://www.mdpi.com/2072-4292/11/7/830}
\BIBentrySTDinterwordspacing

\bibitem{8428417}
X.~Qin, S.~He, X.~Yang, M.~Dehghan, Q.~Qin, and J.~Martin, ``Accurate outline
  extraction of individual building from very high-resolution optical images,''
  \emph{IEEE Geoscience and Remote Sensing Letters}, vol.~15, no.~11, pp.
  1775--1779, 2018.

\bibitem{ronneberger2015u}
O.~Ronneberger, P.~Fischer, and T.~Brox, ``U-net: Convolutional networks for
  biomedical image segmentation,'' in \emph{Medical Image Computing and
  Computer-Assisted Intervention--MICCAI 2015: 18th International Conference,
  Munich, Germany, October 5-9, 2015, Proceedings, Part III 18}.\hskip 1em plus
  0.5em minus 0.4em\relax Springer, 2015, pp. 234--241.

\bibitem{badrinarayanan2017segnet}
V.~Badrinarayanan, A.~Kendall, and R.~Cipolla, ``Segnet: A deep convolutional
  encoder-decoder architecture for image segmentation,'' \emph{IEEE
  transactions on pattern analysis and machine intelligence}, vol.~39, no.~12,
  pp. 2481--2495, 2017.

\bibitem{jegou2017one}
S.~J{\'e}gou, M.~Drozdzal, D.~Vazquez, A.~Romero, and Y.~Bengio, ``The one
  hundred layers tiramisu: Fully convolutional densenets for semantic
  segmentation,'' in \emph{Proceedings of the IEEE conference on computer
  vision and pattern recognition workshops}, 2017, pp. 11--19.

\bibitem{sun2019high}
K.~Sun, Y.~Zhao, B.~Jiang, T.~Cheng, B.~Xiao, D.~Liu, Y.~Mu, X.~Wang, W.~Liu,
  and J.~Wang, ``High-resolution representations for labeling pixels and
  regions,'' \emph{arXiv preprint arXiv:1904.04514}, 2019.

\bibitem{zhu2020map}
Q.~Zhu, C.~Liao, H.~Hu, X.~Mei, and H.~Li, ``Map-net: Multiple attending path
  neural network for building footprint extraction from remote sensed
  imagery,'' \emph{IEEE Transactions on Geoscience and Remote Sensing},
  vol.~59, no.~7, pp. 6169--6181, 2020.

\bibitem{chen2022asf}
J.~Chen, Y.~Jiang, L.~Luo, and W.~Gong, ``Asf-net: Adaptive screening feature
  network for building footprint extraction from remote-sensing images,''
  \emph{IEEE Transactions on Geoscience and Remote Sensing}, vol.~60, pp.
  1--13, 2022.

\bibitem{liu2019net}
H.~Liu, J.~Luo, B.~Huang, X.~Hu, Y.~Sun, Y.~Yang, N.~Xu, and N.~Zhou, ``De-net:
  Deep encoding network for building extraction from high-resolution remote
  sensing imagery,'' \emph{Remote Sensing}, vol.~11, no.~20, p. 2380, 2019.

\bibitem{OpenBuildingsGoogle}
``Open buildings by google,'' Available at
  \url{https://sites.research.google/open-buildings/} (accessed July 20, 2023).

\bibitem{sirko2021continental}
W.~Sirko, S.~Kashubin, M.~Ritter, A.~Annkah, Y.~S.~E. Bouchareb, Y.~Dauphin,
  D.~Keysers, M.~Neumann, M.~Cisse, and J.~Quinn, ``Continental-scale building
  detection from high resolution satellite imagery,'' \emph{arXiv preprint
  arXiv:2107.12283}, 2021.

\bibitem{offsetissue1}
``Why-does-my-gps-data-and-imagery-not-line-up?'' Available at
  \url{https://www.agsgis.com/Advanced-Mobile-Mapping-Series-Why-Does-My-GPS-Data-and-Imagery-Not-Line-Up_b_1062.html}
  (accessed July 20, 2023).

\bibitem{yuan2017learning}
J.~Yuan, ``Learning building extraction in aerial scenes with convolutional
  networks,'' \emph{IEEE transactions on pattern analysis and machine
  intelligence}, vol.~40, no.~11, pp. 2793--2798, 2017.

\end{thebibliography}
\bibliographystyle{IEEEtran}




%

%
%
%




\end{document}